%% file: main.tex
\pdfoutput=1
\documentclass{article}

\usepackage[final]{acl}

\usepackage{multirow} 
\usepackage{pifont}
\usepackage{times}
\usepackage{latexsym}
\usepackage[utf8]{inputenc} 
\usepackage[T1]{fontenc}    
\usepackage{hyperref}       
\usepackage{url}            
\usepackage{booktabs}       
\usepackage{amsfonts}       
\usepackage{nicefrac}       
\usepackage{microtype}      
\usepackage{xcolor}         
\usepackage{caption}
\usepackage{subcaption}
\usepackage{graphicx}
\usepackage{amsmath}
\usepackage{amssymb}
\usepackage{booktabs}
\usepackage{pifont} 
\usepackage{tabularx}
\usepackage{scrextend}
\usepackage{enumitem}
\usepackage{algorithm, algorithmic}
\newtheorem{definition}{Definition}
\usepackage[many]{tcolorbox}        
\definecolor{main}{HTML}{5989cf}    
\definecolor{sub}{HTML}{cde4ff}     

\tcbset{
    sharp corners,
    colback = white,
    before skip = 0.2cm,
    after skip = 0.5cm
}    
\newtcolorbox{boxK}{
    sharpish corners, 
    boxrule = 0pt,
    toprule = 4.5pt, 
    enhanced,
    fuzzy shadow = {0pt}{-2pt}{-0.5pt}{0.5pt}{black!35} 
}

\newcommand{\llama}{\includegraphics[height=1em]{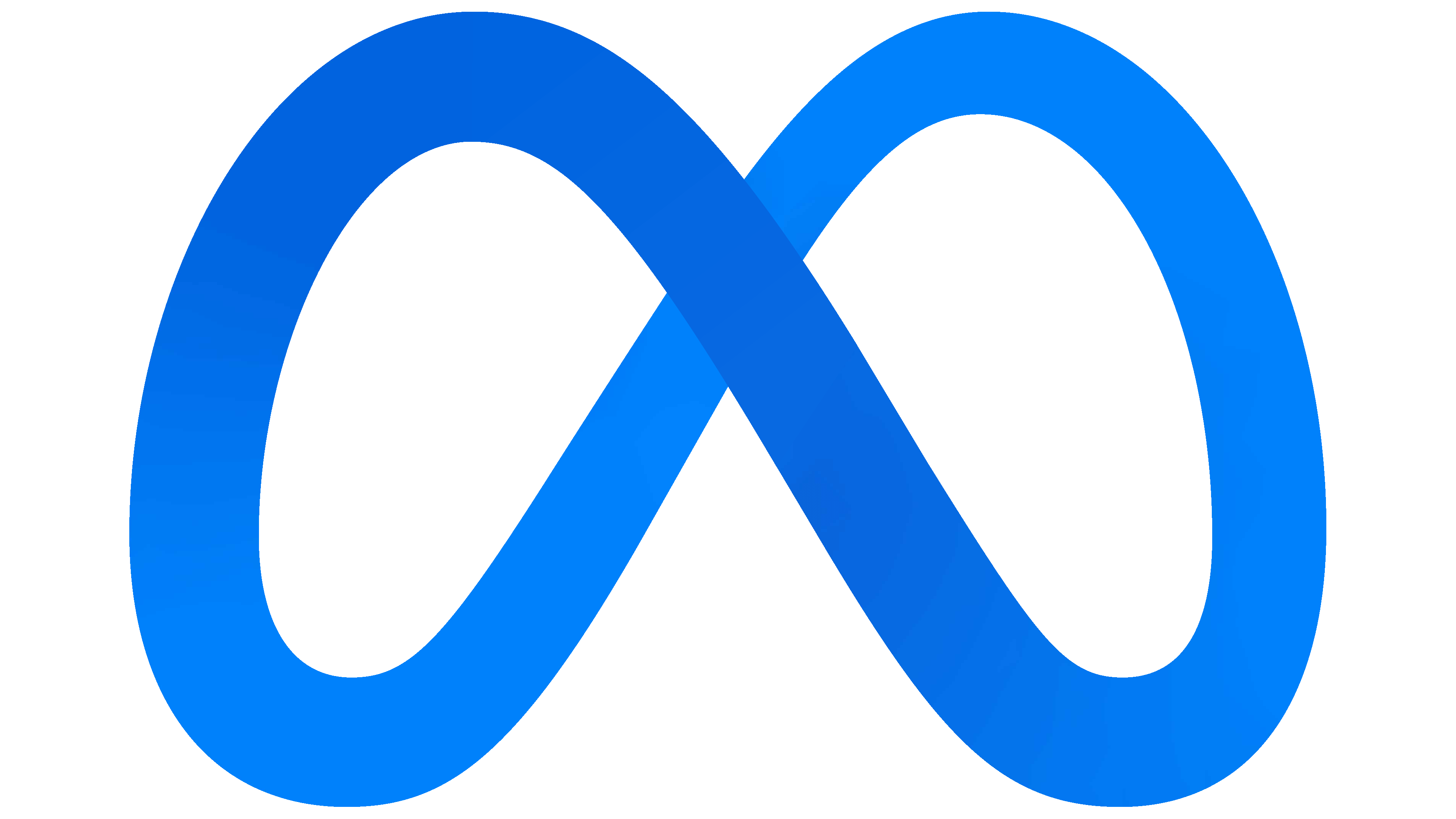}}
\newcommand{\gpt}{\includegraphics[height=1em]{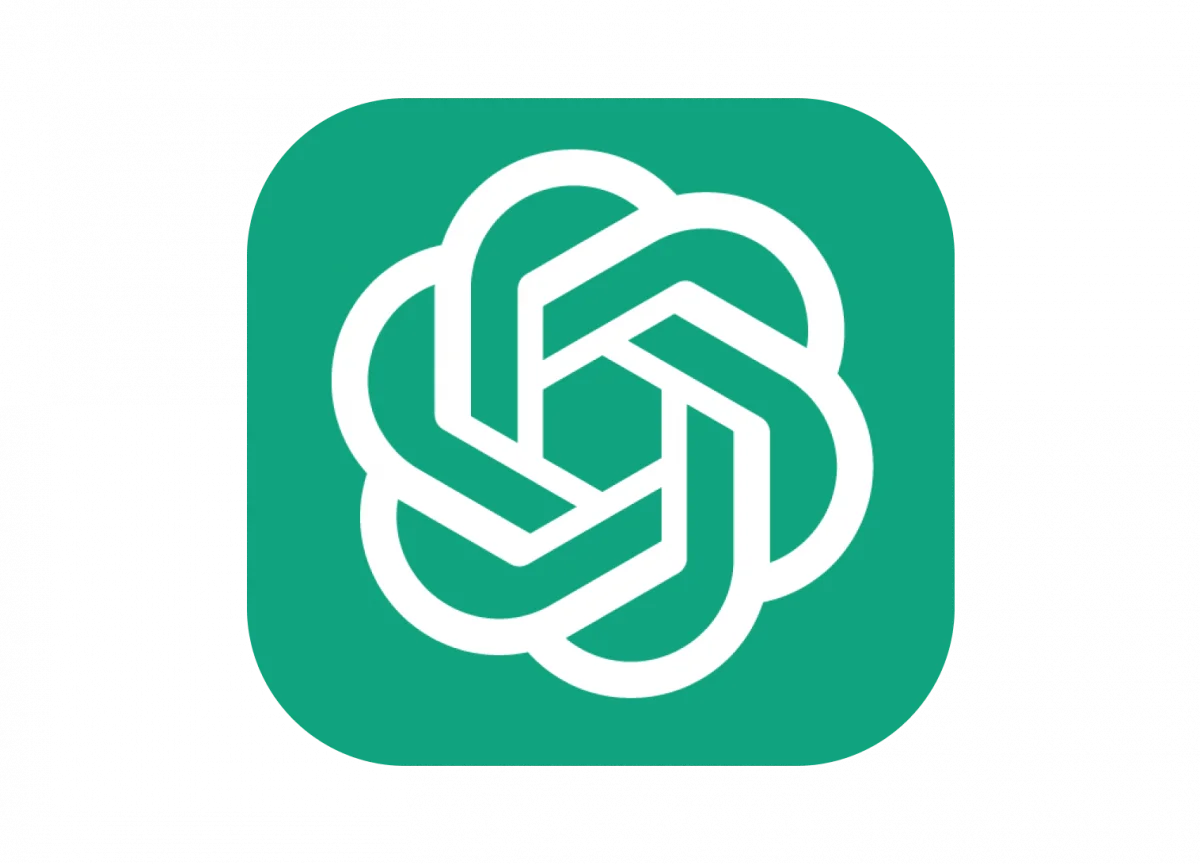}}

\title{Beyond True or False: Retrieval-Augmented\\Hierarchical Analysis of Nuanced Claims}

\input{src/author_info}

\begin{document}

\maketitle

\def\thefootnote{*}\footnotetext{Equal contribution.}\def\thefootnote{\arabic{footnote}}

\input{src/0_abstract}

\input{src/1_introduction}

\input{src/2_related_works}

\input{src/3_method}

\input{src/4_results}

\input{src/5_conclusion}

\bibliography{ref}

\input{src/6_appendix}

\end{document}

%% file: src/author_info.tex
  \author{Priyanka Kargupta\footnote{Equal contribution}, Runchu Tian$^*$, Jiawei Han\\
  Department of Computer Science, University of Illinois at Urbana-Champaign\\
  \texttt{\{pk36, runchut2, hanj\}@illinois.edu} \\
}


%% file: src/0_abstract.tex
\begin{abstract}

Claims made by individuals or entities are oftentimes nuanced and cannot be clearly labeled as entirely ``true'' or false''---as is frequently the case with scientific and political claims. However, a claim (e.g., ``vaccine A is better than vaccine B'') can be dissected into its integral aspects and sub-aspects (e.g., efficacy, safety, distribution), which are individually easier to validate. This enables a more comprehensive, structured response that provides a well-rounded perspective on a given problem while also allowing the reader to prioritize specific angles of interest within the claim (e.g., safety towards children). Thus, we propose \textbf{\textsc{ClaimSpect}}, a retrieval-augmented generation-based framework for automatically \textit{constructing} a hierarchy of aspects typically considered when addressing a claim and \textit{enriching} them with corpus-specific perspectives. This structure hierarchically partitions an input corpus to retrieve relevant segments, which assist in discovering new sub-aspects. Moreover, these segments enable the discovery of varying perspectives towards an aspect of the claim (e.g., support, neutral, or oppose) and their respective prevalence (e.g., ``how many biomedical papers believe vaccine A is more \textit{transportable} than B?''). We apply \textsc{ClaimSpect} to a wide variety of real-world scientific and political claims featured in our constructed dataset, showcasing its robustness and accuracy in deconstructing a nuanced claim and representing perspectives within a corpus. Through real-world case studies and human evaluation, we validate its effectiveness over multiple baselines.
\end{abstract}

%% file: src/1_introduction.tex
\section{Introduction}

\par Scientific and political topics increasingly being consumed in the form of concise, attention-grabbing claims which lack the nuance needed to represent complex realities \cite{vosoughi2018spread, allcott2017social,lazer2018science}. Such claims are frequently oversimplified or confidently stated, despite being valid only under specific conditions or when evaluated from certain perspectives. For instance, a claim like ``vaccine A is better than vaccine B'' may appear straightforward but becomes inherently nuanced when specific aspects, such as efficacy, safety, and distribution logistics, are considered. Moreover, the ambiguous and fragmented nature of information shared on such platforms often allows such claims to be twisted or reframed as ``true'' or ``false'' to support conflicting narratives, complicating the task of verifying their validity \cite{sharma2019combating, pennycook2021psychology}.

\begin{figure}
    \centering
    \includegraphics[width=1\linewidth]{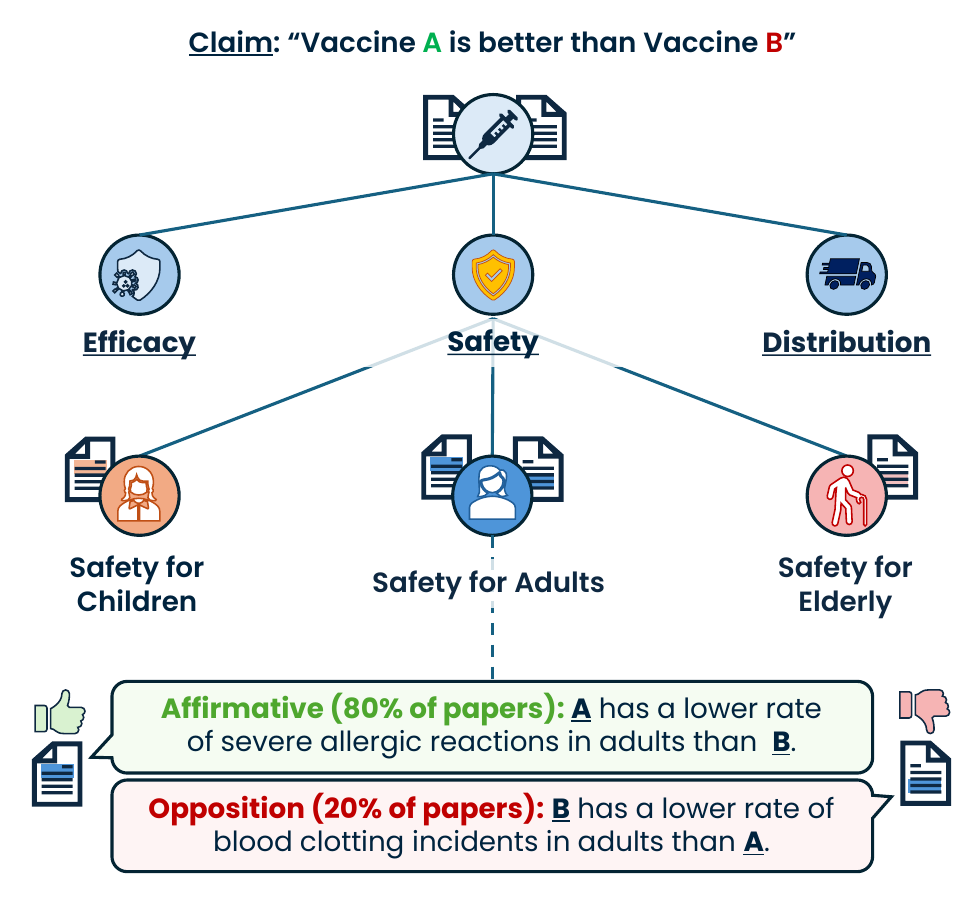}
    \caption{An example hierarchy of a nuanced claim being deconstructed into aspects. Each node is enriched with relevant excerpts, the affirmative/neutral/opposing perspectives, and their respective evidence.}
    \label{fig:example}
\end{figure}

\par Stance detection categorizes textual opinions as supportive, neutral, or opposing relative to a target \cite{mohammad2016semeval}. However, documents---especially those in a scientific domain---often present a \textit{range of stances} across various aspects of a claim. For instance, as illustrated in Figure \ref{fig:example}, a study might find Vaccine A safer for adults than Vaccine B while highlighting its significantly greater logistical challenges for widespread distribution. In this case, the paper supports the claim regarding ``safety for adults'' (note: not ``safety'' in its entirety) but opposes it concerning distribution. This complexity renders stance detection at the document level ineffective for nuanced, multifaceted claims.


\par Fact-checking models often validate claims by retrieving evidence from large corpora or using web-integrated language models \cite{thorne-etal-2018-fever, popat2018declare, zhang-gao-2023-towards}. While some methods now offer varied factuality judgements like ``mostly true'' or ``half-true'' \cite{zhang-gao-2023-towards}, these are less effective in scientific contexts. Especially in evolving areas, fine-grained scientific claims may be \textit{unsubstantiated} due to a \textit{lack of research} or \textit{scientific consensus}, rather than being outright false. This distinction is vital, as it highlights areas needing further exploration. For example, in Figure \ref{fig:example}, relevant paper excerpts mapped to the ``Safety for Adults'' node show that an 80:20 ratio of affirmative to opposing stances towards the sub-aspect claim suggests consensus, whereas a 60:40 ratio or sparse data signals limited research or disagreement. Such insights, crucial for understanding gaps in knowledge, are often \textbf{overlooked by existing fact-checking frameworks}.

\par We address these challenges using \textsc{ClaimSpect}, a framework which systematically deconstructs and analyzes claims by leveraging large language models (LLMs). ClaimSpect hierarchically partitions a claim into a tree of aspects and sub-aspects, enabling structured validation and the discovery of perspectives. This is accomplished by adopting the following principles:

\par{\textbf{\underline{Principle \#1:} Claim trees capture the  multidimensionality inherent in nuanced topics.}} As opposed to considering a single target claim and the full document, we must first determine the relevant aspects discussed within the corpus itself in order to discover more targeted subclaims. However, it is essential to retain the hierarchical nature of such aspects. This is demonstrated in Figure \ref{fig:example}, where certain aspects that are difficult to validate (e.g., ``safety'') can typically be partitioned until they reach ``atomic'' sub-aspects that are more commonly considered (e.g., ``safety for children'', ``safety for adults'', and ``safety for elderly''). Furthermore, these hierarchical relationships are often also reflected in how we naturally navigate formulating our own perspective towards a given topic (either individually or collectively): parse through the existing knowledge on a topic, consider different sub-angles of the problem based on this knowledge, retrieve more sub-angle specific knowledge, develop our opinions accordingly, and aggregate them to a high-level opinion \cite{perony2013enhancing,chen2022expertise}. Thus, this brings us to our next principle.


\par{\textbf{\underline{Principle \#2:} Iterative, discriminative retrieval enhances LLM-based tree construction.}} LLMs have recently shown promise in automatic taxonomy enrichment and expansion, organizing data into hierarchies of categories and subcategories similar to our target aspect hierarchy \cite{shen2024unified,zeng2024codetaxo}. However, these approaches often rely on general knowledge existing within the LLM's pre-training dataset, overlooking corpus-specific insights crucial for (1) uncovering fine-grained sub-aspects prevalent in domain-specific data, and (2) ensuring alignment with the task of determining corpus-wide consensus.  To address this, we leverage retrieval-augmented generation (RAG), which has recently made advances in knowledge-intensive tasks by integrating external corpora or databases into the generation process \cite{lewis2020retrieval,gao2023retrieval}. We introduce an \textit{iterative} RAG approach, which dynamically constructs the aspect hierarchy by retrieving relevant segments for an aspect node, using them to \textit{discover new sub-aspects}. This ensures the taxonomy aligns closely with corpus-specific discussions of claims, aspects, and perspectives.
\par We note that noisy retrieval often hinders reasoning performance \cite{shen-etal-2024-assessing}. In our setting, this may occur when certain retrieved excerpts overlap multiple semantically similar aspect nodes (e.g., ``safety for children'' vs. ``safety for adults''), introducing noise when determining sub-aspects for only one aspect. To mitigate this, we introduce a discriminative ranking mechanism that prioritizes segments discussing a \textit{single} aspect \textit{in-depth}, enhancing sub-aspect discovery and the final aspect hierarchy.

\par{\textbf{\underline{Principle \#3:} Perspectives enrich understanding beyond stance and consensus.}} For each aspect node in the hierarchy, we identify and cluster papers based on their stance (affirmative, neutral, opposing) using hierarchical text classification and stance detection. These clusters reveal not only the presence or absence of consensus but also the key perspectives within each stance. For example, as shown in Figure \ref{fig:example}, the affirmative perspective might highlight Vaccine A’s lower rate of severe allergic reactions in adults, while the opposition focuses on its higher incidence of blood clotting. These perspectives offer transparency, uncover potential research gaps (e.g., if 80\% of the affirmative papers do not address these blood clotting incidents), and provide critical context for framing nuanced claims.

\par Overall, \textbf{\textsc{ClaimSpect}} utilizes a structured approach to deconstruct a nuanced claim into a hierarchy of aspects, targeting a holistic approach considering all aspects which could be used to validate the root claim. The framework comprises the following steps: \textit{\textbf{(1)} aspect-discriminative retrieval, \textbf{(2)} iterative sub-aspect discovery, and \textbf{(3)} classification-based perspective discovery.} Our contributions can be summarized as: 
\begin{itemize}
\itemsep0em
    \item From the best of our knowledge, \textsc{ClaimSpect} is the \textit{first work} to formally deconstruct claims into a hierarchical structure of aspects to determine consensus.
    \item We construct \textbf{two novel datasets} of real-world, scientific and political \textit{nuanced} claims and corresponding corpora.
    \item Through \textbf{experiments and case studies on real-world domains}, we demonstrate that ClaimSpect performs hierarchical consensus analysis significantly more effectively than the baselines.
\end{itemize}

\textbf{Reproducibility:} We provide our dataset and source code\footnote{\url{https://github.com/pkargupta/claimspect}} to facilitate further studies.

%% file: src/2_related_works.tex
\section{Related Works}

\par{\textbf{Fact Checking.}} Fact-checking models \cite{thorne-etal-2018-fever, popat2018declare,atanasova2019automatic,karadzhov-etal-2017-fully} have leveraged external evidence to validate claims, but often treat claims as monolithic statements. Web-integrated methods \cite{zhang-gao-2023-towards,karadzhov-etal-2017-fully} attempt to enrich fact-checking with additional context, but still fail to account for \textit{nuanced} claims that cannot be clearly validated without considering a diverse range of claim sub-aspects and their varying levels of evidence. In contrast, \textsc{ClaimSpect} acknowledges the nuance behind certain claims, utilizing a corpus to help identify the various aspects that would be considered when validating a claim--- enabling a more multi-faceted and interpretable analysis. We note that \textsc{ClaimSpect} does not aim to validate a given claim--- it simply aims to \textbf{\textit{deconstruct}} the claim into a hierarchy of aspects which \textit{could be used to validate it}, \textbf{\textit{posing potential perspectives}} towards the aspect of the claim, \textbf{\textit{grounded}} in the corpus. We also note the adjacent task of evidence retrieval, where existing work explores organizing evidence according to a fixed and flat (non-hierarchical) set of aspects: populations, interventions, and outcomes \cite{wadhwa-etal-2023-redhot}.

\par{\textbf{LLM-Based Taxonomy Generation.}} Recent advances in taxonomy generation \cite{shen2024unified, zeng2024codetaxo,chen2023prompting,zeng2024chain,sun2024large} have demonstrated the potential of large language models for structuring information hierarchically. However, these methods typically rely on static, domain-agnostic knowledge, limiting their adaptability to construct rich, fine-grained taxonomies \cite{sun2024large}. \textsc{ClaimSpect} addresses these limitations through corpus-aware, aspect-discriminative retrieval and iterative sub-aspect discovery, constructing a rich \textit{taxonomy} of aspects that is aligned with a corpus. This allows us to identify the relevant segments to both a given aspect but also a perspective towards that aspect.

\par{\textbf{Stance Detection}} Traditional stance detection \cite{mohammad2016semeval} classifies opinions as supportive, neutral, or opposing towards a target (e.g., claim). However, these approaches typically assign a single stance to an entire document, overlooking the nuanced, aspect-specific stances present within many claims, especially in scientific and political contexts. Recent works \cite{zhang-gao-2023-towards} have introduced more fine-grained judgments (e.g., ``mostly true''), but similar to fact-checking methods, they often fail to capture the multi-faceted nature and \textit{rationale} behind certain stances. By exploiting its constructed aspect hierarchy, \textsc{ClaimSpect} is able to infer viable \textit{supportive}, \textit{neutral}, and \textit{opposing \textbf{perspectives} towards an aspect} and its associated papers.

%% file: src/3_method.tex
\begin{figure*}
    \centering
    \includegraphics[width=1.0\linewidth]{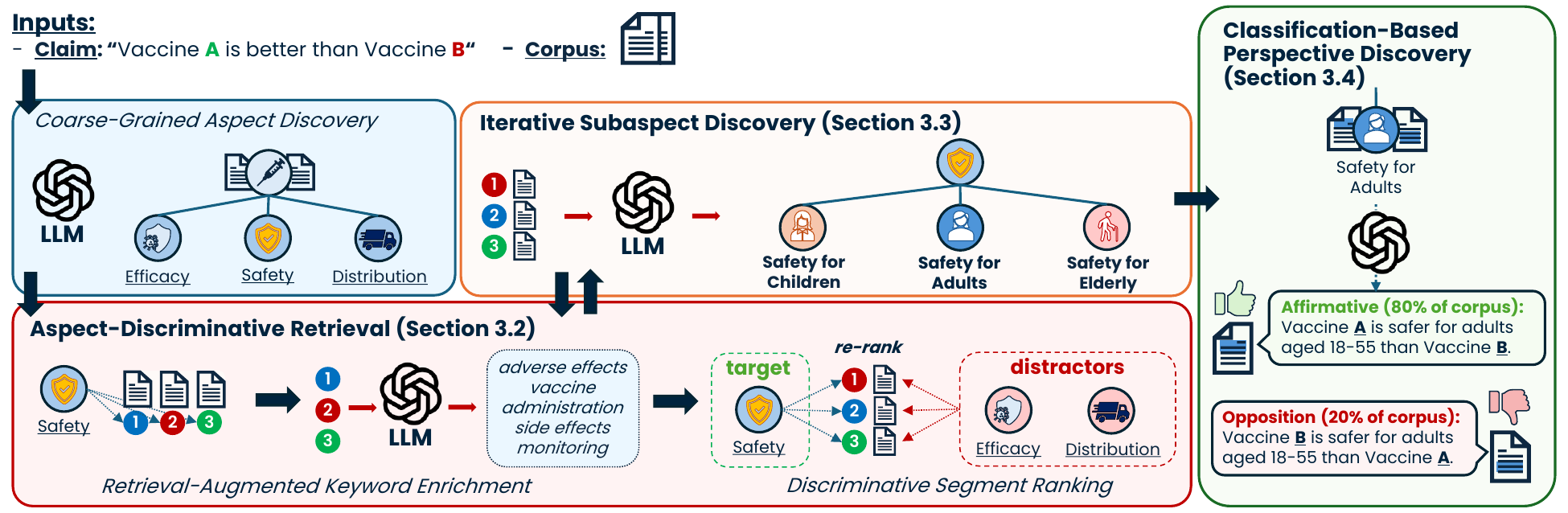}
    \caption{\textsc{ClaimSpect} deconstructs a nuanced claim into a hierarchy of aspects typically considered for validating the claim. We automatically discover the set of perspectives towards each aspect from the corpus.}
    \label{fig:framework}
\end{figure*}

\section{Methodology}
\par Illustrated in Figure \ref{fig:framework}, \textsc{ClaimSpect} consists of the following steps: \textit{\textbf{(1)} aspect-discriminative retrieval, \textbf{(2)} iterative sub-aspect discovery, and \textbf{(3)} classification-based perspective discovery.}

\subsection{Preliminaries}
\label{subsection: preliminaries}

\subsubsection{Task Definition}
\label{sec: task_definition}
\par We assume that as input, the user provides a claim $t_0$ (e.g.,``Vaccine A is better than Vaccine B'') and a corpus $D$. In order to better reflect real-world settings, we \textit{do not} assume that each document $d \in D$ is relevant to $t_0$.
\begin{definition}[\textsc{Claim}]
    A statement or assertion that expresses a position, which may require validation or scrutiny. It often encapsulates multiple dimensions that contribute to its overall truthfulness or validity.
\end{definition}
\begin{definition}[\textsc{Aspect}]
    A specific component or dimension of a claim that can be independently analyzed or evaluated.
\end{definition}
ClaimSpect aims to output a hierarchy of aspects $T$, where each aspect node (e.g., ``safety'') within the hierarchy can be considered as a descendant subclaim $t_i$ of the root user-specified claim, $t_0$ (e.g., ``A is a safer vaccine than B''). In other words, each aspect node $t_i$ should reflect a relevant aspect that is important to consider when evaluating the root claim $t_0$.

\subsubsection{Document Preprocessing}
\par For each $d \in D$, we assume we have its full textual content (e.g., a full scientific paper). In order to have smaller, context-preserving units of text for our framework to retrieve, we segment each $d$ into chunks using the widely-recognized text segmentation method, C99 \cite{choi-2000-advances}. This method labels sentences with matching tags if they pertain to the same topical group, which assists with retaining consecutive discussion of an aspect to a single segment.

\subsubsection{Initial Coarse-Grained Aspect Discovery}
\label{sec: coarse-aspects}
\par Given our weak supervision setting, where only the root claim $t_0$ is provided, we first generate reliable, coarse-grained aspects to guide the retrieval-augmented hierarchy construction. These aspects are typically common-sense and do not require domain expertise to identify. Preliminary experiments confirm that LLMs can reliably identify them based on their expansive background knowledge alone. Thus, we prompt an LLM to generate coarse-grained aspects $t_i^{0} \in T^0$ (e.g., efficacy, safety, and distribution in Figure \ref{fig:example}) that will serve as the children of $t_0 \in T$. For each aspect $t_i^0$, the model outputs its label, significance to $t_0$, and a list of $n=10$ relevant keywords. This initial subtree forms the foundation of our framework. The full prompt is in Appendix \ref{appendix: coarse-grained prompt}.

\subsection{Aspect-Discriminative Retrieval}
\label{sec:disc_retrieval}
\par In order to construct a rich, coarse-to-fine aspect hierarchy that is \textit{aligned with} the corpus, we must identify similarly rich reference material \textit{from our} corpus. In general, noisy retrieval often hinders reasoning performance \cite{shen-etal-2024-assessing}, which may negatively impact discovering subaspects of a given node. Thus, in order to discover each subaspect $t_j^i$ of an aspect node $t_i$, we must determine which segments $S_i$ from our corpus $D$ discuss $t_i$. However, not all segments are equally informative for discovering subaspects.

\par Specifically, a high-quality, \textbf{\textit{discriminative}} segment $s_i$ for node $t_i$ contains the following features: \textbf{(1)} $s_i$ discusses $t_i$ in \textit{depth} and \textbf{(2)} $s_i$ \textit{does not} discuss $t_i$'s siblings in \textit{breadth} or \textit{depth}. For instance, in Figure \ref{fig:example}, a segment regarding the side effects observed within a clinical trial of Vaccine A and B on both children and adults \textit{discusses ``safety'' in more depth} than if it only mentioned children. Furthermore, for discovering subaspects of ``safety for children'', a segment which independently discusses the safety for \textit{both} children and adults would introduce additional noise into the subaspect generation process. Overall, it is important to rank these segments such that we select a set which \textbf{\textit{minimizes} the noise} we introduce into the retrieval-augmented discovery of subaspects, while \textbf{\textit{maximizing} the \textit{number} of subaspects} which we can discover. We formalize our discriminative ranking mechanism in the sections below:

\subsubsection{Retrieval-Augmented Keyword Enrichment}
\label{sec: keyword-enrich}
\par In order to determine whether a segment discusses an aspect $t_i$ \textit{in depth}, we must first further enrich our understanding of $t_i$. We propose performing a retrieval-augmented keyword-based enrichment of $t_i$, where each keyword is likely to occur within segments relevant to $t_i$ and, thus, reflects either explicitly or implicitly the subaspects of $t_i$. For example, for the ``\textit{efficacy}'' aspect, the corresponding keywords are: \textit{neutralization, immune stimulation, post-dose antibody response, and waning immunity}. First, we use a retrieval embedding model to select the top-$n$ segments (based on cosine-similarity) from the entire corpus that are relevant to a $t_i$-specific query (its root, name, description, and keywords from Section \ref{sec: coarse-aspects}):

\begin{addmargin}[1em]{0em}
\textit{\textbf{Claim}: [$t_0$]; \textbf{Aspect}: [$t_i$]: [generated description of $t_i$]; \textbf{Aspect Keywords}: [generated keywords of $t_i$].}
\end{addmargin}
\vspace{0.15cm}
\par We provide these initial top $n$ segments in addition to the root claim $t_0$, the aspect label $t_i$, and its description to the LLM in-context to identify $2k$ keywords. Given the same information and these keywords, we then merge similar or duplicate terms, while filtering irrelevant terms-- explicitly prompting the model to provide solely $k$ keywords. This set of terms $w \in W_i; |W_i| = k$, grounds our discriminative segment ranking for node $t_i$. We provide these two prompts in Appendix \ref{appendix: keyword generation}.

\subsubsection{Discriminative Segment Ranking}
\label{sec: rank_segments}
\par In order to determine the most discriminative segments $S_i$ for aspect node $t_i$, we first collect an initial large pool of segments using the same retrieval embedding-based method as Section \ref{sec: keyword-enrich}. Our subsequent goal is to rank a segment $s \in S_i$ based on its discriminativeness:
\begin{itemize}[leftmargin=*]
\itemsep0em
    \item \textbf{Target Score:} Reward $s$ based on its likelihood to contain all relevant subaspects $t_j^i$ of $t_i$.
    \item \textbf{Distractor Score:} Penalize $s$ based on the degree and depth of other sibling aspects that it discusses.
\end{itemize}

\par We assume that $t_i$s' keywords $W_i$ implicitly and/or explicitly reflect many of its subaspects. Thus, we use them to approximate the depth of an aspect-specific discussion. We convert each keyword $w_i$ into a descriptive query: \textit{``[$w_i$] with respect to [all ancestor nodes of $w_i$]''}. By integrating the ancestors into the query, we influence the retention of $t_i$'s hierarchical context; for example, we specifically \textit{reward} a segment if it discusses ``the safety of \underline{Vaccine A and B}'', as opposed to merely ``safety''. We embed each keyword query $emb(w \in W_i)$ using the retrieval embedding model, in addition to embedding each segment $emb(s) \in S_i$.

\par More formally, we are given an aspect node $t_i^h$, which is a child of parent node $t_h$ and sibling node of $t_j \in T_{\neq i}^{h}$. We are also provided with a segment embedding $emb(s) \in S_i$, all keyword query embeddings of $t_i^h$, $emb(w) \in W_i$, and all sibling keyword query embeddings, $emb(w) \in W_{\neq i}^h$. We compute the discriminative rank based on the following:
\begin{definition}[\textsc{Target Score}]
A segment $s_i$ is rewarded based on a weighted average ($H$) of its degree of similarity to all keywords $w \in W_i$, implying a deeper discussion of node $t_i$ and its subaspects.
\end{definition}
\begin{equation}
\small
    \label{tag:positive_score}
    \begin{split}
        \textbf{p}(s_i, W_i) &= H\bigg(\bigg[ \textbf{sim}\big(emb(s_i), emb(w)\big) \mid w \in W_i\bigg]\bigg), \\
        &\text{where } H(X) = \frac{\sum_{r=1}^{|X|} \frac{1}{r}x_r}{\sum_{r=1}^{|X|} \frac{1}{r}}
    \end{split}
\end{equation}

\par We compute a weighted average based on Zipf's Law \cite{powers-1998-applications}, where a word indexed at the $r$-th position will have a weight of $1/r$. This weighted average of the segment-keyword similarities is based on the assumption that the model will implicitly generate the keywords from most to least significant-- in other words, we weight the first term $w_1 \in W_i$ the highest, while weighing $w_{|X|}$ the lowest. For example, if $s_i$ had similarities of $[0.9, 0, 0]$ to $W_i = \{w_1,w_2,w_3\}$, then $\textbf{p}(s_i, W_i) = 0.5363$. On the other hand, if the similarities were $[0.7, 0.8, 0.7]$, $\textbf{p}(s_i, W_i) = 0.7272$. Overall, the target score will indicate a segment's discussion \textit{depth} of aspect node $t_i$-- \textit{how many} keywords it aligns with and \textit{to what degree}.

\begin{definition}[\textsc{Distractor Score}]
A segment $s_i$ is penalized based on the \textit{breadth} and \textit{depth} of siblings discussed. The breadth is indicated by the mean target scoring between $s_i$ and each $W_j$ of $t_j \in T_{\neq i}^{h}$. The depth is indicated by the max target scoring between $s_i$ and each $W_j$ of $t_j \in T_{\neq i}^{h}$.
\end{definition}
\begin{equation}
\small
    \label{tag:negative_score}
    \begin{split}
        \textbf{n}(s_i, T_{\neq i}^{h}) &= 0.5 \times \bigg(\frac{1}{|T_{\neq i}^{h}|} \sum_{j=1}^{|T_{\neq i}^{h}|} p(s_i, W_j) \bigg) \\
        &+ 0.5 \times \bigg( max_{j = \big[1, |T_{\neq i}^{h}|\big]}\big( p(s_i, W_j) \big) \bigg)
    \end{split}
\end{equation}

\par We utilize the target and distractor scores to compute our overall discriminativeness score, which weighs the proximity between a segment and its target aspect, relative to its overall and individual proximity to its distractor, sibling aspects.

\begin{definition}[\textsc{Discriminativeness Score}]
A segment $s_i$ is rewarded based on a weighted average ($H$) of its degree of similarity to all keywords $w \in W_i$, while being penalized based on the \textit{breadth} and \textit{depth} of siblings discussed.
\end{definition}
\begin{equation}
\small
    \label{tag:disc_score}
        \textbf{d}(s_i, W^{h}) = \frac{\beta \times p(s_i, W_i^h)}{\gamma \times n(s_i, T_{\neq i}^{h})}
\end{equation}

\par In Equation \ref{tag:disc_score}, \textbf{d}$(s_i, W^{h})$ grows proportional to the target score, while falling proportional to the distractor score. We include the $\beta$ and $\gamma$ scaling factors for each in case users would like to customize their degree of reward or penalty. Ultimately, we rank each segment $s \in S_i$ based on its discriminativeness score, taking the top-$k$ segments which feature the richest discussion of target aspect $t_i$ in order to discover its subaspects.

\subsection{Iterative Subaspect Discovery}
\label{sec:subaspect_discovery}
\par In order to expand our aspect hierarchy, we iteratively exploit our aspect-discriminative retrieval as knowledge which grounds the LLM's subaspect discovery. Given the aspect node $t_i$, its description, its corresponding discriminative segments $S_i$, and the root claim $t_0$, we prompt the model to determine a set of at minimum two and at maximum $k$ subaspects for aspect $t_0$. We provide this prompt in Appendix \ref{appendix: subaspect_prompt}.
\begin{definition}[\textsc{Subaspect}]
    A more granular component of a parent aspect $t_i$ that further refines $t_i$'s evaluation and would be considered when specifically addressing the root claim $t_0$.
\end{definition}
Each subaspect is represented in the same manner specified in Section \ref{sec: coarse-aspects}: its label, description, and keywords. We continue constructing our aspect hierarchy in a top-down fashion, as detailed in Algorithm \ref{algorithm: subaspect-pseudocode}.

\input{src/algorithm}

\par Ultimately, the output of Algorithm \ref{algorithm: subaspect-pseudocode} is our final aspect hierarchy, serving as the basis for our consensus determination and perspective discovery process.

\subsection{Classification-Based Perspective Discovery}
\label{sec:perspective_disc}

\par With the aspect hierarchy constructed, we must identify the \textit{complete} set of corpus segments that (1) pertain to the root claim $t_0$ and (2) align with an aspect node in hierarchy $T$. Pinpointing papers discussing aspect node $t_i$ allows us to infer their \textit{perspective} on $t_i$ and assess the \textit{presence} and \textit{extent of consensus}. However, as noted in Section \ref{sec: task_definition}, we cannot assume all corpus segments are relevant to the root claim—-an assumption made in LLM-based taxonomy-guided hierarchical classification works \cite{zhang2024teleclass}. Thus, we must first filter out claim-irrelevant segments.

\par{\textbf{Filtering.}} A naive approach determines segment relevance per node via in-context prompting, but this scales poorly. Instead, we frame relevance filtering as a \textit{binary search} problem, identifying the relevance-irrelevance boundary. Specifically, we embed the claim label $t_0$ ($emb(t_0)$) and each child aspect $t^0_i \in T^0$ ($emb($``\texttt{[aspect\_label]} with respect to [$t_0$]''$)$), computing the claim representation as:

\begin{equation}
    \textbf{c}_0 = \frac{1}{2}\bigg( emb(t_0) +  \frac{\sum_{i=1}^{|T^0|}emb(t^0_i)}{|T^0|}\bigg)
\end{equation}

We rank the encoded segments by cosine-similarity to $\textbf{c}_0$ and use binary search to find the index $r$ where fewer than $\delta\%$ of segments in a $\pm n$ window are relevant. This rank $r$ serves as our threshold, filtering out lower-ranked segments and retaining only those relevant to $t_0$ ($S'_0$). This optimization significantly reduces the quantity of relevance judgments necessary; the relevancy prompt is in Appendix \ref{appendix: relevancy_prompt}.

\par{\textbf{Hierarchical Text Classification.}} With $S'_0$ and $T$, we apply \textit{taxonomy-guided hierarchical classification} to determine $S'_i$ for each aspect node $t_i \in T$. Since our focus is retrieval-guided aspect hierarchy construction rather than classification, we adopt a recent LLM-based hierarchical classification model \cite{zhang2024teleclass}, which enriches taxonomy nodes (e.g., adding keywords) to support its top-down classification of $S'_i$ to $t_i$.


\par{\textbf{Perspective \& Consensus Discovery.}} The final step of our pipeline is to determine the primary perspectives $P_i = \{a_i, o_i\}$ towards each aspect node $t_i$, where $a_i$ is the overarching supportive perspective and $o_i$ is the opposing perspective. We also seek to identify the papers which hold each of these perspectives ($D_i = D_i^{\text{supp}} \space \cup \space D_i^{\text{opp}} \space \cup D_i^{\text{neutral}}$), accounting for papers which do not hold any clear perspective towards $t_i$.

\begin{definition}[\textsc{Perspective}]
    A descriptive viewpoint expressed toward a specific aspect $t_i$ of claim $t_0$ in the form of an implicit or explicit stance towards $t_i$ (e.g., support, neutral, or oppose) and optionally, a rationale.
\end{definition}

\par We do not assume that $D_i^{\text{supp}}, D_i^{\text{opp}}\text{, and } D_i^{\text{neutral}}$ are non-overlapping, as they may have multiple segments indicating different perspectives. For example, a segment $s'_i \in S'_i$ mapped to ``Safety for Elders'' may discuss a clinical trial showing increased anaphylactic shock in older patients when taking Vaccine A. However, another segment from the same paper may also note severe hives from Vaccine B. Thus, we allow for such flexibility.

\par Recent studies have shown LLMs demonstrate powerful abilities in stance detection \cite{zhang2024llm, lan2024stance}. Consequently, in order to discover these perspectives, we prompt the model to first determine the stance of each segment $s'_i \in S'_i$:
\begin{itemize}[leftmargin=*]
\itemsep-0.5em
    \item \textbf{Supports Claim}: $s'_i$ either implicitly or explicitly indicates that the \textbf{\textit{claim is true}} with respect to $t_i$.
    \item \textbf{Neutral to Claim}: $s'_i$ is relevant to the claim and aspect, but \textbf{\textit{does not indicate}} whether the claim is true with respect to $t_i$.
    \item \textbf{Opposes Claim}: $s'_i$ either implicitly or explicitly indicates that the \textbf{\textit{claim is false}} with respect to $t_i$. 
\end{itemize}

\par This forms the segment sets: $S_i^{\text{supp}}$, $S_i^{\text{neutral}}$, and $S_i^{\text{opp}}$. We ask the model to summarize the perspective (stance and rationale) of each segment set: $P_i$. Both prompts are provided in Appendix \ref{appendix: perspective_prompt}.  Since we retain the original paper source of each segment, we are able to construct $D_i$ for each node $t_i$. This indicates consensus; for instance, how many papers in $D$ held perspective $p_i^{\text{supp}}$ towards aspect $t_i$. \textbf{As our final output, we have the aspect hierarchy $T$, the set of perspectives $P_i$, and their corresponding papers $D_i$.}

%% file: src/algorithm.tex
\begin{algorithm}[h!]
\small
    \caption{Iterative Subaspect Discovery}
    \label{algorithm: subaspect-pseudocode}
    \begin{algorithmic}[1]
        \REQUIRE Root Claim $t_0$, Corpus $D$, max\_depth=$l$
        \STATE $T = $ initialize\_tree($t_0$) \COMMENT{$T$.depth $= 0$}
        \STATE $t_i^0 \in T^0 \leftarrow$ coarse\_grained\_aspects($t_0$) \COMMENT{Section \ref{sec: coarse-aspects}}
        \STATE $q$ = queue($T^0$)
        \WHILE{$len(q) > 0 \text{ and } T.\text{depth} \leq l$}
            \STATE $t_i \leftarrow pop(q)$
            \STATE enrich\_node($t_0, t_i,D$) \COMMENT{Section \ref{sec: keyword-enrich}}
            \STATE $S_i \leftarrow $ rank\_segments($t_0, t_i, D$) \COMMENT{Section \ref{sec: rank_segments}}
            \STATE $t_j^i \in T^i\leftarrow$ subaspect\_discovery($t_0, t_i, S_i$)
            \STATE $q$.append($T^i$)
        \ENDWHILE
        \RETURN $T$
    \end{algorithmic}
\end{algorithm}

%% file: src/4_results.tex
\section{Experimental Design}
\par We explore \textbf{\textsc{ClaimSpect}}'s performance on an open-source model, \texttt{Llama-3.1-8B-Instruct} (\llama). We sample from the top 1\% of the tokens and set the temperature based on the nature of the given task (same setting across all samples); we include these settings in Appendix \ref{temp}. We set the maximum depth of the aspect hierarchy to $l = 3$.

\subsection{Dataset}
\par In order to evaluate \textsc{ClaimSpect}'s abilities to deconstruct \textbf{\textit{nuanced claims}} into a hierarchy of aspects and identify their corresponding perspectives, we construct \textbf{two novel, large-scale datasets} specific to our task, applied to both political (\textbf{World Relations}) and scientific (\textbf{Biomedical}) domains. To construct this dataset, we first manually collect $\sim$50 reference materials discussing (1) security-related international conflicts, and (2) biomedical safety-related studies. Then, we used \texttt{GPT-4o}~\citep{openai2024gpt4ocard} to generate nuanced claims based on these materials. Subsequently, we used the Semantic Scholar API~\citep{semanticscholarapi} to collect meta information relevant literature based on these claims. Then, based on this meta information, we filtered the collected literature and retrieved the full texts. This way, for each claim, we obtained a corresponding literature repository. We show the statistics of each of these datasets in Table \ref{tab:dataset}.  More details about the dataset construction can be found in Appendix~\ref{appendix: Dataset Construction}, including a human study for validating the quality of the generated claims and their associated papers in Appendix \ref{sec:claimvalidation}.


\begin{table}[h]
\small
\renewcommand{\arraystretch}{1.1}
\centering
\setlength{\tabcolsep}{5pt}
\begin{tabular}{l c c c}
    \toprule
    \textbf{Dataset} & \textbf{Claims} & \textbf{Papers} & \textbf{Segments} \\
    \midrule
    \textbf{World Relations} &  140  & 9,525 & 1,081,241\\
    \textbf{Biomedical} &  50  & 3,719 & 428,833 \\
    \midrule
    \textbf{Total}     & 190 & 13,244 & 1,510,074 \\
    \bottomrule
\end{tabular}
\caption{\textit{\# of claims, papers, and segments per dataset.}}
\label{tab:dataset}
\end{table}

\input{src/tables}

\subsection{Baselines}
\par Our primary motivation for \textsc{ClaimSpect} is to demonstrate its capabilities of deconstructing a nuanced claim into an aspect hierarchy and identifying corresponding perspectives. However, no existing methods tackle this novel task. Consequently, we choose to implement and compare our method with both \textit{RAG-driven} and \textit{LLM-only} approaches, detailed below. We run each baseline using both Llama (\llama) and GPT-4o-mini (\gpt):
\begin{enumerate}
    \item \textbf{RAG-Based:} Given a claim and definition of an aspect hierarchy, we use the claim as a query to retrieve relevant documents. We then provide the documents in-context when prompt the LLM to generate the aspect hierarchy.
    \item \textbf{Iterative RAG-Based:} Given the claim, the definition of an aspect hierarchy, and the name/description of the current node $t_i$, we construct a detailed query to retrieve node-specific relevant documents. We then provide these documents in-context to prompt the LLM for generating the children subaspects $t^i_j$ of aspect $t_i$.
\end{enumerate}
\par We also conduct an ablation study, \textbf{\textit{No Discriminative}} (No Disc), where we remove discriminative ranking and instead replace it with a semantic similarity-based ranking. For this, we compute the semantic similarity between each segment and our $t_i$-specific query from Section \ref{sec: keyword-enrich}.

\subsection{Evaluation Metrics}
\par We design a thorough automatic evaluation suite using \texttt{GPT-4o-mini} to determine the quality of our generated taxonomies, using both node-level and taxonomy-level metrics. For each judgment, we ask the LLM to provide additional rationalization:

\begin{itemize}[leftmargin=*]
    \itemsep-0.25em
    \item \textbf{(\textit{Node-Wise}) Node Relevance:} For each aspect node $t_i$ and its respective path within the hierarchy, what is its relevance to the claim $t_0$? Scored 0/1.
    \item \textbf{(\textit{Node-Wise}) Path Granularity:} Does the path to node $t_i$ preserve the hierarchical relationships between its entities (is each child $t_j^i$ more specific than the parent $t_i$)? Scored 0/1.
    \item \textbf{(\textit{Level-Wise}) Sibling Granularity:} For each set of siblings $T^i$ within the hierarchy, does the overall set reflect the same level of specificity relative to their parent aspect $t_i$? Scored from 1 to 4 (all different $\rightarrow$ some $\rightarrow$ most $\rightarrow$ all same).
    \item \textbf{(\textit{Node-Wise}) Uniqueness:} Does the aspect node $t_i$ have other overlapping nodes within the hierarchy $T$? Scored 0/1.
    \item \textbf{\textit{(Node-Wise)} Segment Quality:} How many segments $s \in S'_i$ are relevant to the claim $t_0$ and aspect $t_i$? We compute the average proportion of relevant segments per node.
\end{itemize}

In addition to automatically evaluating our aspect hierarchy, we also conduct a supplementary human evaluation on 50 perspectives and their sampled segments, which \textsc{ClaimSpect} identifies from the corpus (Section \ref{sec:eval_perspective}).

\begin{table*}[ht!]
    \centering
    \footnotesize
    \caption{Pairwise comparisons between all methods for each dataset. Each value is the \textit{percentage} of samples within each dataset where the method is considered better. \textbf{\textit{E-Tie}} denotes Explicit Tie; \textbf{\textit{I-Tie}} denotes Implicit Tie.}
    \renewcommand{\arraystretch}{1.5}
    \begin{tabular}{|l|c c c c|c c c c|}
        \hline
        \multirow{2}{*}{\textbf{Method Pair (A vs. B)}} & \multicolumn{4}{c|}{\textbf{World Relations}} & \multicolumn{4}{c|}{\textbf{Biomedical}} \\
        \cline{2-9}
         & \textit{\textbf{A Wins}} & \textit{\textbf{B Wins}} & \textit{\textbf{E-Tie}} & \textit{\textbf{I-Tie}} & \textit{\textbf{A Wins}} & \textit{\textbf{B Wins}} & \textit{\textbf{E-Tie}} & \textit{\textbf{I-Tie}} \\
        \hline
        Zero-Shot \llama vs RAG \llama & 0.00 & 33.06 & 0.00 & \textbf{66.94} & 2.22 & 22.22 & 0.00 & \textbf{75.55} \\
        Zero-Shot \llama vs \textbf{\textsc{ClaimSpect}} \llama & 0.00 & \textbf{97.58} & 0.00 & 2.42 & 0.00 & \textbf{95.55} & 2.22 & 2.22 \\
        RAG \llama vs \textbf{\textsc{ClaimSpect}} \llama & 0.81 & \textbf{90.32} & 0.00 & 8.87 & 0.00 & \textbf{95.55} & 0.00 & 4.44 \\
        \textbf{\textit{No Disc}} \llama vs \textbf{\textsc{ClaimSpect}} \llama & 21.43 & 30.00 & 0.00 & \textbf{48.57} & 24.00 & 28.00 & 0.00 & \textbf{48.00} \\
        \midrule
        Zero-Shot \gpt vs RAG \llama & 0.00 & 36.00 & 0.00 & \textbf{64.00} & 0.71 & 47.14 & 0.00 & \textbf{52.14} \\
        Zero-Shot \gpt vs \textbf{\textsc{ClaimSpect}} \llama & 0.00 & \textbf{98.00} & 0.00 & 2.00 & 0.00 & \textbf{96.43} & 0.00 & 3.57 \\
        RAG \gpt vs \textbf{\textsc{ClaimSpect}} \llama & 0.00 & \textbf{90.00} & 0.00 & 10.00 & 7.14 & \textbf{72.14} & 0.71 & 20.00 \\
        \hline
    \end{tabular}
    \label{tab:pairwise_comparisons_updated}
\end{table*}

\begin{figure*}[h!]
    \centering
    \includegraphics[width=1.0\textwidth]{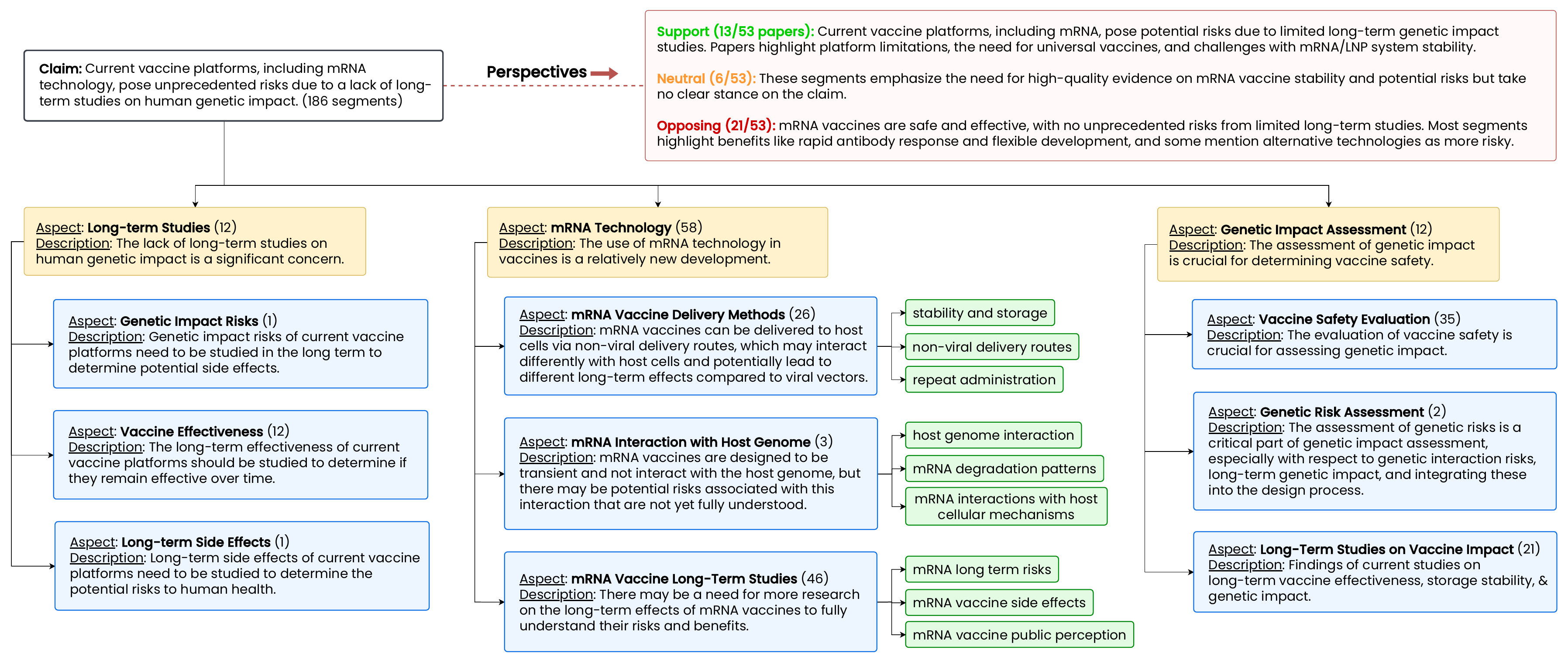}
    \caption{A constructed Biomedical aspect hierarchy. All nodes and their \# of segments from levels 1-2 are included; a subset of the third level is highlighted. The \# of papers mapped to each perspective is provided in parentheses.}
    \label{fig:case_study}
\end{figure*}

\section{Experimental Results}
\subsection{\textbf{Overall Performance \& Analysis}}
\par Tables \ref{tab:quant_results}-\ref{tab:pairwise_comparisons_updated} demonstrate several key advantages of \textsc{ClaimSpect} over the baselines across various node and level-wise metrics for both the \textit{World Relations} and \textit{Biomedical} datasets. \textsc{ClaimSpect} is able to strongly enforce the hierarchical structure of the generated aspect hierarchy while preserving relevance to the corpus. Below, we present our core findings and insights. We also provide a breakdown of ClaimSpect's computational efficiency in Appendix \ref{app:efficiency}. Finally, we additionally conducted a human-automatic evaluation agreement study in Appendix \ref{app:alignment}.

\par{\textbf{\textsc{ClaimSpect} excels in \underline{granular} aspect discovery.}} As shown in Table \ref{tab:quant_results}, \textsc{ClaimSpect} significantly outperforms the baselines in metrics associated with node-level structure, particularly outperforming Iterative RAG \llama by $72.6\%$ and $63.51\%$ in preserving hierarchical relationships (path granularity) and by $44.48\%$ and $26.61\%$ in maintaining uniform sibling-level specificity (sibling granularity) for both datasets respectively. This demonstrates the method's ability to \textit{retrieve and organize aspects at targeted levels of granularity}. These gains are similarly observed with the GPT-based baselines, despite relying on a closed-source model. We attribute this gain to ClaimSpect's iterative subaspect discovery (Section \ref{sec:subaspect_discovery}) being integrated with its \textit{aspect-discriminative retrieval mechanism} (Section \ref{sec:disc_retrieval}), where the pool of segments grounding the subaspect discovery is iteratively updated based on the given aspect node. We can see that the \textit{No Disc} ablation does experience some loss in granularity quality. It is important to note that \textit{No Disc} does experience competitive and, at times, better performance; this is likely due to it considering more segments, which may or may not discuss multiple aspects. In contrast, the baseline methods retrieve broader, less focused segments, reducing their ability to discover fine-grained sub-aspects. Overall, this demonstrates that ClaimSpect is able to \textbf{\textit{deconstruct a claim into a well-structured hierarchy of aspects}}.

\par{\textbf{\textsc{ClaimSpect} constructs a \underline{rich} aspect hierarchy while preserving \underline{relevance}.}} In Table \ref{tab:quant_results}, we observe that ClaimSpect's constructed hierarchy features nodes that are $14.40\%$ and $11.23\%$ more unique than the top baseline on each dataset, respectively. This indicates that ClaimSpect's hierarchies are \textbf{\textit{richer}} in aspect quality, experiencing less overlap between aspects across the tree and supported by an increase in segment quality. Despite this significant boost in uniqueness, ClaimSpect only experiences a $3.35\%$ and $0.386\%$  drop from the top baseline in aspect node relevance for the World Relations and Biomedical datasets, respectively. This highlights the strength of ClaimSpect's retrieval-augmented keyword enrichment and aspect-discriminative retrieval (Sections \ref{sec: keyword-enrich} and \ref{sec:disc_retrieval}), which prioritize segments that \textit{thoroughly discuss a single aspect} rather than \textit{shallow} descriptions of \textit{multiple} aspects. This allows us to \textbf{\textit{discover a richer set of unique and relevant subaspects at each level, throughout the hierarchy}}.

\par{\textbf{\textsc{ClaimSpect} is overwhelmingly \underline{preferred} over baselines.}} Table \ref{tab:pairwise_comparisons_updated} presents pairwise comparisons between \textsc{ClaimSpect} and the baseline methods. These comparisons are judged by an LLM that is shown the aspect hierarchy outputs of methods $A$ and $B$ in both possible orders: $A$ vs. $B$ and $B$ vs. $A$. The LLM may (1) prefer either method $A$ or $B$, (2) declare an explicit tie (\textbf{\textit{E-Tie}}), or (3) indicate an implicit tie (\textbf{\textit{I-Tie}}), which occurs when the preferred method changes depending on the order of presentation (e.g., $A$ wins in $A$ vs. $B$, but $B$ wins in $B$ vs. $A$) \cite{shi2024judging}.

\par Across both datasets, \textsc{ClaimSpect} exhibits a clear advantage, being preferred $92.95\%$ of the time, with a $6.69\%$ average inconsistency rate across all settings and datasets. Specifically, when compared with Zero-Shot \llama, \textsc{ClaimSpect} is judged superior in \textbf{$97.58\%$} and \textbf{$95.55\%$} of cases for World Relations and Biomedical datasets, respectively. Even against RAG \gpt, \textsc{ClaimSpect} outperforms in \textbf{$90.00\%$} and \textbf{$72.14\%$} of samples. This is a stark contrast from the lack of strong preference between the baselines themselves, indicated by the 64.66\% average implicit tie rate--- implying that there is no obvious qualitative preference between the two. Finally, we show that ClaimSpect and No Disc are similarly preferred, with ClaimSpect preferred slightly more often. Overall, these results validate that \textsc{ClaimSpect} \textbf{\textit{constructs significantly more meaningful aspect hierarchies relevant to the claim}}.

\subsection{Perspective Discovery Analysis}
\label{sec:eval_perspective}

\par{\textbf{\textsc{ClaimSpect} identifies nuanced, corpus-specific perspectives.}} We showcase a qualitative analysis of a nuanced claim's aspect hierarchy, highlighting certain subtrees and the root node's extracted perspectives, in Fig.~\ref{fig:case_study}. We observe each coarse-grained aspect (yellow nodes) well represents the various angles of the root claim that would be considered in validating it: what long-term vaccine studies currently exist, what is the current mRNA technology, and how is genetic impact currently assessed? We see that the path-specific dependencies are reflected within the descriptions of each aspect (e.g., ``\textit{mRNA Interaction with Host Genome}'' involves both mRNA technology \textit{and} potential genetic impact risks). Furthermore, these hierarchical relationships and claim relevance are preserved even in the final layer of the hierarchy (e.g., ``\textit{mRNA Interaction with Host Genome}'' $\rightarrow$ ``\textit{mRNA degradation patterns}''). Finally, we see that the perspectives mapped to the root node are informative, providing justification behind each stance. Note that ClaimSpect maps segments to each perspective, allowing us to identify the original paper sources and ultimately provide a \textbf{\textit{corpus-specific estimate of the consensus}}. Overall, this deconstructed view of the claim provides a means to identify \textbf{\textit{which and to what degree certain aspects have been explored}} (e.g., \textit{mRNA Technology} has been more explored within the corpus compared to \textit{Genetic Impact Assessment}).

\begin{table}[h]
\small
\renewcommand{\arraystretch}{1.1}
\centering
\setlength{\tabcolsep}{5pt}
\begin{tabular}{l c c}
    \toprule
    \textbf{$k$} & \textbf{World Relations} & \textbf{Biomedical}\\
    \midrule
    $5$ & 72\% (50) & 72\% (50)\\
    $10$ & 80\% (20) & 82\% (20) \\
    $15$ & 85\% (19) & 89\% (9) \\
    \bottomrule
\end{tabular}
\caption{Human validation on the percentage of perspectives discovered by \textsc{ClaimSpect} which are grounded in at least one of $k$ associated segments. $k =$ \# of segments considered. We provide the number of samples for each setting in parenthesis.}
\label{tab:human_eval}
\end{table}

\par{\textbf{Human annotators validate the grounding of discovered perspectives.} To assess the validity of the perspectives discovered by \textsc{ClaimSpect}, we apply human evaluation to evaluate whether these perspectives are effectively grounded in the corpus. We randomly sampled perspectives along with their associated $k$ segments (each aspect node has three ass from the generated results across two datasets. The evaluation metric used was \textit{whether at least one segment in $k$ could provide grounding background knowledge for the corresponding perspective}. As shown in Table~\ref{tab:human_eval}, we found that the vast majority of cases (85\% and 89\% for each dataset respectively) are supported by specific literature segments. Furthermore, we can see that the support rate steadily increases as we retrieve more segments that are mapped to the perspective. \textbf{\textit{This shows the perspectives identified by \textsc{ClaimSpect} are largely supported by the corpus}}.}

%% file: src/tables.tex
\begin{table*}[h]
    \small
    \centering
    \caption{Comparison between ClaimSpect and all baselines. Sibling granularity (\textbf{\textit{Sib}}) is normalized; all others are scaled by 100. Since Iterative Zero-Shot is not grounded with a corpus, there are no associated segments to each node. Thus, we omit its segment relevance scores (\textbf{\textit{Seg}}). We \textbf{bold} the top score and \underline{underline} the second-highest.}
    \renewcommand{\arraystretch}{1.5}
    \begin{tabular}{|l|c c c c c|c c c c c|}
        \hline
        \multirow{2}{*}{\textbf{Method}} & \multicolumn{5}{c|}{\textbf{{World Relations}}} & \multicolumn{5}{c|}{\textbf{Biomedical}} \\
        \cline{2-11}
         & \textit{\textbf{Rel}} & \textit{\textbf{Path}} & \textit{\textbf{Sib}} & \textit{\textbf{Unique}} & \textit{\textbf{Seg}} & \textit{\textbf{Rel}} & \textit{\textbf{Path}} & \textit{\textbf{Sib}} & \textit{\textbf{Unique}} & \textit{\textbf{Seg}} \\
        \hline
        Iterative Zero-Shot \llama & 97.85 & 41.94 & 58.01 & 72.96 & --- & \textbf{98.33} & 44.44 & 57.04 & 77.17 & --- \\
        
        Iterative RAG \llama & 97.18 & 45.34 & 59.01 & 74.25 & 42.79 & 97.14 & 45.93 & 59.08 & 76.17 & 27.11 \\
        
        Iterative Zero-Shot \gpt & \underline{98.60} & 42.88 & 64.04 & 76.01 & --- & 97.89 & 41.56 & 62.09 & 77.55 & --- \\
        
        Iterative RAG \gpt & 97.40 & 52.30 & 66.45 & 76.59 & \underline{46.93} & 94.37 & 50.07 & 64.21 & 77.05 & \underline{31.82} \\
        
        \midrule
        \textbf{\textsc{ClaimSpect}} \llama & 95.30 & \underline{78.24} & \textbf{85.26} & \textbf{87.62} & 43.23 & \underline{97.95} & \underline{75.10} & \textbf{74.80} & \underline{86.26} & 27.39 \\
        
        \textit{\textsc{ClaimSpect} - No Disc} \llama & \textbf{99.00} & \textbf{79.75} & \underline{82.64} & \underline{85.43} & \textbf{49.47} & 96.07 & \textbf{76.26} & \underline{74.39} & \textbf{87.69} & \textbf{39.03} \\
        
        \hline
    \end{tabular}
    \label{tab:quant_results}
\end{table*}

%% file: src/5_conclusion.tex
\section{Conclusion}

Our work introduces \textbf{\textsc{ClaimSpect}}, a novel framework for deconstructing nuanced claims into a hierarchy of corpus-specific aspects and perspectives. By integrating iterative, aspect-discriminative retrieval with hierarchical sub-aspect discovery and perspective clustering, \textsc{ClaimSpect} provides a structured, comprehensive view of complex claims. Our experiments on two novel, large-scale datasets demonstrate that \textsc{ClaimSpect} constructs rich, corpus-aligned aspect hierarchies that are enriched with diverse and informative perspectives. This highlights its effectiveness as a scalable and adaptable method for nuanced claim analysis across domains.

\section{Limitations \& Future Work}

\par The primary contribution of \textbf{\textsc{ClaimSpect}} is our retrieval-augmented framework for constructing an aspect hierarchy relevant for validating a nuanced claim. In order to demonstrate the hierarchy's potential, we apply it to the task of perspective discovery, involving (1) identifying which segments from the corpus are \textit{relevant to a given aspect node}, (2) determining the \textit{stance} (or lack thereof) of the segment towards the claim and aspect, and (3) discovering the potential \textit{perspective} of each of the stance-based segment clusters. Consequently, this step \textit{relies heavily upon an existing hierarchical classification} model \cite{zhang2024teleclass}, as we do not claim novelty with respect to classification. Similarly, our classification-based perspective discovery (Section \ref{sec:perspective_disc}) is reliant on the LLM's fine-grained stance detection abilities--- although prior work \cite{zhang2024llm,lan2024stance} has shown precedence for its capabilities. Thus, the performance of the hierarchical classification and stance detection serves as a bottleneck to our method's performance. For example, if the LLM-based stance detection has a \textit{high recall but low precision} for detecting segments which \textit{support} the aspect of claim, then the method may \textit{overestimate} the consensus behind a certain perspective within the corpus. Likewise, if the detection has a high precision but lower recall, it may \textit{underestimate the consensus}. Nonetheless, our work aims to, overall, motivate the need to structure the aspects of certain nuanced claims \textit{before diving straight into their validation.}

\par Hierarchically analyzing nuanced claims opens up doors to many new avenues of research. First, \textsc{ClaimSpect} can be integrated with more systematic and/or tool-integrated fact validation systems, in an effort to build a more robust fact-checking system. Furthermore, \textsc{ClaimSpect} can be applied to more targeted retrieval or question answering tasks where a question, similar to a nuanced claim, cannot easily be answered and may benefit from a more structured output (similar to an aspect hierarchy).

\section{Acknowledgements}
\par This work was supported by the National Science Foundation Graduate Research Fellowship. The work was also supported in part by the BRIES Program No. HR0011-24-3-0325. This research used the DeltaAI advanced computing and data resource, which is supported by the National Science Foundation (award OAC 2320345) and the State of Illinois. DeltaAI is a joint effort of the University of Illinois at Urbana-Champaign and its National Center for Supercomputing Applications. We thank Peter Bautista, Spencer Lynch, and Svitlana Volkova from Aptima, Inc. for their discussions on our work. We also thank Mihir Kavishwar for early ideation discussions.

%% file: src/6_appendix.tex
\clearpage

\appendix

\section{Human Annotator Background and Human-Automatic Alignment for Evaluation}
\label{app:alignment}
We perform a human evaluation study to show the alignment between humans and \texttt{GPT-4o-mini} for evaluation. We use two human evaluators to evaluate randomly sampled cases for each of our proposed metrics: node relevance, path granularity, sibling granularity, uniqueness and segment quality. Our human evaluators are two volunteer graduate researchers (one PhD student and Masters student), with one having a background in Biology + NLP (critical for our biomedical task evaluation). We used the following instructions to help guide them in the evaluation task, with an initial ``training'' period where the evaluators could familiarize themselves with the task (e.g., aspect taxonomies and their expected hierarchical relationships) and discuss examples with one another, but the full evaluation period was conducted independently:
\begin{enumerate}
    \item \textbf{General Instruction}: \textit{Claims made by individuals or entities are often nuanced and cannot always be strictly categorized as entirely 'true' or 'false', particularly in scientific and political contexts. Instead, a claim can be broken down into its core aspects and sub-aspects, which are easier to evaluate individually.}
    \item \textbf{Node Relevance}: \textit{Given the claim: [claim], decide whether this path from the aspect tree is relevant to the analysis of the claim: [path]} $\rightarrow$ relevant or irrelevant?
    \item \textbf{Path Granularity}: \textit{Given the claim: [claim], decide whether this path from the aspect tree has good granularity: [path]. Check whether the child node is a more specific subaspect of the parent node.} $\rightarrow$ granular or non-granular?
    \item \textbf{Sibling Granularity}: \textit{Given the claim: [claim]', decide whether these siblings from parent node [parent] have good granularity.} $\rightarrow$ <proportion of granular siblings>
    \item \textbf{Uniqueness}: \textit{Normally, we want the aspects and sub-aspects to be unique in the taxonomy. Given the claim: [claim], count how many nodes in this taxonomy are largely overlapping or almost equivalent.} $\rightarrow$ <total count of overlapping nodes>
    \item \textbf{Segment Quality}: \textit{Given the claim: [claim], evaluate the quality of these segments for aspect [aspect node label]. [list of mapped segments]} $\rightarrow$ <total count of relevant segments to aspect node>
\end{enumerate}

\par Since ClaimSpect has consistently high scores on ``Node Relevance'' and ``Uniqueness'' (not very many nodes are irrelevant or overlapping across the entire taxonomy), the scores from both the LLM and evaluators have very low variance. This greatly limits both Cohen’s 
 and Intraclass Correlation Coefficient (high instability). Nonetheless, the evaluators have a $100\%$ and $96.97\%$ agreement rate respectively. We choose Cohen's $\kappa$ or ICC based on the ordinal/continuous versus categorical nature of the metric.

 \begin{enumerate}
     \item The weighted $\kappa$ for Path Granularity is $0.62$ $\rightarrow$ \textbf{substantial agreement}
     \item The $ICC1k$ for Sibling Granularity is $0.7806$ $\rightarrow$ \textbf{good reliability}
     \item The $ICC2k$ for Segment Quality is $0.7578$ $\rightarrow$ \textbf{good reliability}
 \end{enumerate}

\par We also show the agreement rate below on 100 different samples across all metrics in Table \ref{tab:alignment}.

\begin{table}[h]
\centering
\scriptsize
\begin{tabular}{lccccc}
\toprule
& \textbf{Rel} & \textbf{Path} & \textbf{Sib} & \textbf{Unique} & \textbf{Seg} \\
\midrule
\textbf{Agreement Rate} & 100\% & 85\% & 87.5\% & 96.97\% & 82\% \\
\bottomrule
\end{tabular}
\caption{Human-Automatic Agreement rates across different metrics.}
\label{tab:alignment}
\end{table}

\par We can see that the LLM evaluation and human evaluators have a high degree of alignment. We note that segment quality features the lowest alignment (albeit still a relatively high rate), likely due to the more fine-grained text understanding abilities required for verifying segment alignment to the parent node (where oftentimes, segments can be quite semantically dissimilar or only discuss a “sub-aspect” of the node). Nonetheless, through these results, we can see that **our auto-evaluation is reliable**.

\section{Prompt Template}

In this section, we present the prompts used in different modules of \textsc{ClaimSpect}.

\subsection{Coarse-Grained Aspect Discovery}
\label{appendix: coarse-grained prompt}
This is the prompt used to generate coarse-grained aspects for the root claim, including their labels, description, and relevant keywords to structure the initial retrieval-augmented hierarchy.

\begin{boxK}

\textbf{Prompt}
\vspace{0.3em}

For the topic, \{topic\}, output the list of up to \{k\} aspects in JSON format.

\end{boxK}

\subsection{Retrieval-Augmented Keyword Enrichment}
\label{appendix: keyword generation}

Following are the prompts used for retrieval-augmented keyword enrichment, instructing the LLM to refine and filter aspect-specific keywords for improved segment ranking.

\begin{boxK}

\textbf{Prompt (Extraction)}
\vspace{0.3em}

The claim is: \{claim\}. You are analyzing it with a focus on the aspect \{aspect\_name\}. The aspect, \{aspect\_name\}, can be described as the following: \{aspect\_description\}

Please extract at most \{2*max\_keyword\_num\} keywords related to the aspect \{aspect\_name\} from the following documents: 
\{contents\}
Ensure that the extracted keywords are diverse, specific, and highly relevant to the given aspect. Only output the keywords and seperate them with comma.
Your output should be in JSON format.
\end{boxK}

\begin{boxK}

\textbf{Prompt (Filtering)}
\vspace{0.3em}

Our claim is '\{claim\}'. With respective to the target aspect '\{aspect\_name\}', identify \{min\_keyword\_num\} to \{max\_keyword\_num\} relevant keywords from the provided list: \{keyword\_candidates\}.

\{aspect\_name\}: \{aspect\_description\}

Merge terms with similar meanings, exclude relatively irrelevant ones, and output only the final keywords separated by commas.

Your output should be in JSON format.

\end{boxK}

\subsection{Iterative Subaspect Discovery}
\label{appendix: subaspect_prompt}
Following is the prompt used to iteratively guide the LLM in discovering and expanding subaspects for each aspect node based on discriminative retrieval and root claim context.

\begin{boxK}

\textbf{Prompt}

Output the list of up to \{k\} subaspects of parent aspect \{aspect\} that would be considered when evaluating the claim, \{topic\}.
claim: \{topic\}
parent\_aspect: \{aspect\}; \{aspect\_description\}
path\_to\_parent\_aspect: \{aspect\_path\}
Provide your output in the following JSON format.

\end{boxK}

\subsection{Relevance Filtering}
Following is the prompt used for relevance filtering, leveraging binary search on cosine-similarity rankings to efficiently identify and retain only the most relevant segments for each aspect.

\label{appendix: relevancy_prompt}

\begin{boxK}

\textbf{Prompt}

I am currently analyzing a claim based on a segment from the literature from several different aspects.
The segment is: \{segment\}
The claim is: \{claim\}
The aspects are: \{aspects\}
Please help me determine whether this segment is related to the claim so that I can analyze this claim based on it from at least one of these aspects. Your output should be 'Yes' or 'No' in JSON format.
\end{boxK}

\subsection{Perspective Discovery}
\label{appendix: perspective_prompt}

Following are prompts used to for determining segment stances (support, neutral, or oppose) and summarizing perspectives, including rationales, for each aspect.

\begin{boxK}

\textbf{Prompt}
\vspace{0.3em}

You are a stance detector, which determines the stance that a segment from a scientific paper has towards an aspect of a specific claim. Oftentimes, scientific papers do not provide explicit, outright stances, so your job is to figure out what stance the data or statement that they are presenting implies.
Segment: \{segment.content\}

What is the segment's stance specifically with respect to \{aspect\_name\} for if \{claim\}? \{aspect\_name\} can be described as \{aspect\_description\}.
Claim: \{claim\}
Aspect to consider: \{aspect\_name\}: \{aspect\_description\}
Path to aspect: \{aspect\_path\}
    
Your stance options are the following:
- supports\_claim: The segment either implicitly or explicitly indicates that claim is true specific to the given aspect.
- neutral\_to\_claim: The segment is relevant to the claim and aspect, but does not indicate whether the claim is true specific to the given aspect.
- opposes\_claim: The segment either implicitly or explicitly indicates that the claim is false specific to the given aspect.
- irrelevant\_to\_claim: The segment does not contain relevant information on the claim and the aspect.

\end{boxK}

\section{Generation Settings}
\label{temp}

This section details the temperature values used in various stages of our process and their respective roles.

\subsection{Overview of Temperature Settings}

\begin{itemize}
    \item \textbf{Coarse-Grained Aspect Discovery} (0.3): Used to generate high-level aspects related to the claim. A lower temperature ensures structured and deterministic output.
    \item \textbf{Subaspect Discovery} (0.7): Used for identifying subaspects from ranked segments. A higher temperature allows for more diversity while maintaining coherence.
    \item \textbf{OpenAI Chat Models} (GPT-4o~\citep{openai2024gpt4ocard}, GPT-4o-mini~\citep{openai2024gpt4ocard}) (0.3): Applied in various stages where GPT-4o models are used (e.g., aspect generation, classification), ensuring consistent responses.
    \item \textbf{Subaspect Discovery (Aspect Ranking and Retrieval)} (0.7): Used when extracting subaspects from ranked segments to balance creativity with relevance.
\end{itemize}

\subsection{General Trends}

\begin{itemize}
    \item \textbf{Lower temperature} (0.3) is used for structured and deterministic tasks such as \textit{aspect generation and classification}.
    \item \textbf{Higher temperature}
    (0.7) is applied to \textit{subaspect discovery}, where diversity and exploration are beneficial.
\end{itemize}

\section{Dataset Construction}
\label{appendix: Dataset Construction}

To evaluate the effectiveness of \textsc{ClaimSpect}, our nuanced claims analysis, we constructed two datasets covering two key domains: \textbf{political (World Relations)} and \textbf{scientific (Biomedical)}. The dataset construction process consists of the following steps:

\subsection{Manual Seed Collection}
We begin by manually collecting a set of seed claims from reliable sources such as Google Scholar~\citep{GoogleScholar} and PubMed~\citep{PubMed}. Specifically, we collect material from 7 papers in the World Relations domain and 50 papers in the Biomedical domain. These initial materials serve as a context or specific topics for generating nuanced claims.

\subsection{Nuanced Claims Generation}
Using the literature collected in the previous step and definition of nuanced claims as context, we prompt \texttt{GPT-4o}~\citep{openai2024gpt4ocard} to generate nuanced claims related to the topics within these papers. To ensure diversity in claim perspectives, we employ two sets of prompts: one for generating claims that align with the perspectives in the literature and another for generating claims that diverge from them. The specific prompts used are detailed below.

\vspace{0.5em}
\begin{boxK}
\textbf{Positive Claim Generation Prompt}
\vspace{0.3em}

Scientific or political claims are often nuanced and multifaceted, rarely lending themselves to simple “yes” or “no” answers. To answer such questions effectively, claims must be broken into specific aspects for in-depth analysis, with evidence drawn from relevant scientific literature. We are currently studying such claims using this corpus: 

\{context\}

Task: Generate 10 nuanced and diverse claims based on this corpus. The claims should adhere to the following criteria:

	1.	Diversity: The claims should be sufficiently varied: they should involve diverse sub-topics in the context
    
	2.	Complexity: The claims should be complex and controversial (and not necessarity true), requiring multi-aspect analysis rather than simplistic treatment. Avoid overly straightforward or simplistic claims.
    
	3.	Research Feasibility: The claims should not be too specific and should pertain to topics with a likely body of existing literature to support evidence-based exploration.
    
    4.  Concision: The claims should be concise and focused in one short sentence.
    
    5.  Completeness: The claims should be complete and not require additional context to understand.

Output: Provide the claims as a list.

\end{boxK}

\begin{boxK}
\textbf{Negative Claim Generation Prompt}
\vspace{0.3em}

Scientific or political claims are often nuanced and multifaceted, rarely lending themselves to simple “yes” or “no” answers. To answer such questions effectively, claims must be broken into specific aspects for in-depth analysis, with evidence drawn from relevant scientific literature. We are currently studying such claims using this corpus: 

\{context\}

Task: Generate 10 nuanced and diverse claims based on this corpus. The claims should adhere to the following criteria:

	1.	Diversity: The claims should be sufficiently varied: they should involve diverse sub-topics in the context
    
	2.	Complexity: The claims should be complex and controversial (and not necessarity true), requiring multi-aspect analysis rather than simplistic treatment. Avoid overly straightforward or simplistic claims.
    
	3.	Research Feasibility: The claims should not be too specific and should pertain to topics with a likely body of existing literature to support evidence-based exploration.
    
    4.  Concision: The claims should be concise and focused in one short sentence.
    
    5.  Completeness: The claims should be complete and not require additional context to understand.

    \textbf{6.	The claims should be against the point of view in the context.}
    
Output: Provide the claims as a list.

\end{boxK}
\vspace{1em}
We find that the generated nuanced claims are of high quality. They are content-rich, specific, and difficult to classify as simply true or false, aligning well with our task requirements. Below are some example claims from our datasets.
\vspace{1em}

\begin{boxK}
\textbf{Claims for World Relations}
\vspace{0.3em}

1. International collaborations under the Global Nuclear Security Program prioritize geopolitical alliances over immediate nuclear threat reduction.

\vspace{0.3em}

2. Counteracting WMDs through international partnerships creates dependency and may hinder national self-sufficiency in threat reduction capabilities.

\vspace{0.3em}

3. The effectiveness of the biological threat reduction component is questionable given the rise and global spread of emerging biological threats.
\end{boxK}

\begin{boxK}
\textbf{Claims for Biomedical Domain}
\vspace{0.3em}

1. COVID-19 vaccine safety evaluations are compromised by inconsistent application of evidence standards across different data sources like RCTs and VAERS.

\vspace{0.3em}

2. The rigid adherence to optimized distribution plans might inhibit the flexibility needed to respond to unforeseen disruptions in the vaccine supply chain.

\vspace{0.3em}

3. Keeping manufacturing costs secret is essential for protecting proprietary processes and innovations in the pharmaceutical industry.
\end{boxK}

\subsection{Meta Information Collection}
To support the corpus-based analysis of each claim, we retrieve relevant literature using the Semantic Scholar API~\citep{semanticscholarapi}.

Since our claims are highly nuanced and involve multiple concepts, directly searching for claims themselves does not yield useful matches based on literature titles and abstracts. To address this, we first perform keyword extraction for each claim. We then use the extracted keywords to query the Semantic Scholar API and retrieve up to 1000 related literature entries for each claim.

\subsection{Filtering and Full-Text Collection}
After obtaining the literature metadata, we first filter out entries with missing fields and retain the top 100 most relevant papers based on relevance. We then utilize the provided PDF URLs to download the full-text of the selected literature and convert them into plain text with pdftotext~\citep{palmer2024pdftotext}. As a result, we obtain a comprehensive textual literature repository for each claim, ensuring a rich contextual foundation for further analysis.

This structured approach ensures a robust dataset suitable for nuanced claims analysis across the domains.

\subsection{Human Validation of Generated Claims and Assigned Papers}
\label{sec:claimvalidation}
\par We conducted a human evaluation study for validating 40 total claims---20 on each dataset. We define the following binary criteria for claim validation:
\begin{enumerate}
    \item \textbf{Nuanced}: Is the claim obviously true or false?
    \item \textbf{Relevant}: Is the claim relevant to the topic (biomedical/world relations)?
    \item \textbf{Corpus-Aligned@k}: At least $k$ papers are relevant to the claim within the corpus (this is computed on $k=5$ and $k=10$)
\end{enumerate}

We show the validation results in Table \ref{tab:dataset_evaluation}, which demonstrates the nuanced nature of the generated claims, their relevancy, and the presence of $k$ papers aligned to each claim.

\begin{table}[h]
\small
\centering
\begin{tabular}{lcc}
\toprule
\textbf{Metric} & \textbf{World Relations} & \textbf{Biomedical} \\
\midrule
Nuanced              & 0.9  & 1.0 \\
Relevant             & 1.0  & 1.0 \\
Corpus-Aligned@5     & 0.95 & 0.8 \\
Corpus-Aligned@10    & 0.65 & 0.65 \\
\bottomrule
\end{tabular}
\caption{Human validation of claim quality across datasets.}
\label{tab:dataset_evaluation}
\end{table}

\section{Computational Efficiency}
\label{app:efficiency}
\par We specify the components of \textsc{ClaimSpect}'s framework and their corresponding computational efficiency across the entire pipeline below. We consider the number of nodes within a full aspect hierarchy as $n$ and the total number of segments within the corpus as $S$. We additionally provide rough time estimates based on an average sample.

\begin{itemize}
    \item \textbf{Coarse-Grained Aspect Discovery (Section \ref{sec: coarse-aspects})}
    \begin{itemize}
        \item A single LLM call: $O(1)$
    \end{itemize}

    \item \textbf{Aspect-Discriminative Retrieval (Section \ref{sec:disc_retrieval})}
    \begin{itemize}
        \item \textit{Retrieval-Augmented Keyword Enrichment (Section \ref{sec: keyword-enrich})}
        \begin{itemize}
            \item Embedding the segments using the retrieval model takes the most amount of time ($O(S)$), but this can be computed offline given a knowledge base. Retrieval itself is quite efficient since it is embedding-based, and we use cosine-similarity to determine relevance (an efficient computation, especially in high-dimensional scenarios).
            \item Enrich each node: $O(N) \rightarrow $10 seconds per node
        \end{itemize}
        
        \item \textit{Discriminative Segment Ranking (Section \ref{sec: rank_segments})}
        \begin{itemize}
            \item We only compute the ranking on the top-100 segments, so this operation's efficiency is constant: $O(1)$
            \item The target score and distractor score computation scales according to the number of aspects throughout the tree (since their \# of associated keywords is constant): $O(N) \rightarrow $6 seconds for each node
        \end{itemize}
    \end{itemize}

    \item \textbf{Iterative Subaspect Discovery (Section \ref{sec:subaspect_discovery})}
    \begin{itemize}
        \item A single prompt per aspect node in hierarchy: $O(N)$
    \end{itemize}

    \item \textbf{Classification-Based Perspective Discovery (Section \ref{sec:perspective_disc})}
    \begin{itemize}
        \item As mentioned in lines 442--446, ``we reframe relevance filtering as a binary search problem'' intentionally to optimize the efficiency of this module. Thus, instead of sequentially determining claim relevancy for each segment, we sort the segments (which have more than 500 characters) and use binary search ($O(\log S)$) to find the boundary of relevance-irrelevance. This leads to $O(S \log S)$ efficiency due to the sorting function $\rightarrow$ $5$ seconds during runtime, due to Python optimizations.
        \begin{itemize}
            \item Perspective Discovery involves prompting the LLM for each filtered segment’s stance: $O(S) \rightarrow$ 3 minutes
        \end{itemize}
    \end{itemize}
\end{itemize}

We specifically use \texttt{vLLM} to optimize our LLM batched generation. To construct a claim with 39 nodes and a max depth of 3, it takes approximately \textbf{20 minutes} to run on two NVIDIA RTX A6000s. We can see that in total, the core framework operations take \textbf{13 minutes and 42 seconds}, with the remaining time dedicated to embedding computations (which can be done offline).